\documentclass[10pt,twocolumn,letterpaper]{article}
\usepackage{cvpr}
\usepackage{times}
\usepackage{epsfig}
\usepackage{graphicx}
\usepackage{amsmath}
\usepackage{amssymb}
\usepackage{multirow}
\usepackage{graphicx}
\usepackage{xcolor}
\usepackage[ruled,vlined]{algorithm2e}
\usepackage{subcaption}
\usepackage{adjustbox}
\usepackage[breaklinks=true,bookmarks=false]{hyperref}

\newcommand*\samethanks[1][\value{footnote}]{\footnotemark[#1]}

\cvprfinalcopy 

\def\cvprPaperID{****} 

\begin{document}
	
	\title{Improving Domain Generalization by Learning without Forgetting: \\ Application in Retail Checkout}
	\author{
		\stepcounter{footnote}Thuy C. Nguyen\thanks{equal contribution}~~~~~~
		Nam LH. Phan\samethanks~~~~~~
		Son T. Nguyen~~~~~~
		\\
		CyberCore AI\\
		{\small \{thuy.nguyen, nam.phan, son.nguyen\}@cybercore.co.jp
	}}
	\maketitle

	\begin{abstract}	
		Designing an automatic checkout system for retail stores at the human level accuracy is challenging due to similar appearance products and their various poses. This paper addresses the problem by proposing a method with a two-stage pipeline. The first stage detects class-agnostic items, and the second one is dedicated to classify product categories. We also track the objects across video frames to avoid duplicated counting. One major challenge is the domain gap because the models are trained on synthetic data but tested on the real images. To reduce the error gap, we adopt domain generalization methods for the first-stage detector. In addition, model ensemble is used to enhance the robustness of the 2nd-stage classifier. The method is evaluated on the AI City challenge 2022 -- Track 4 and gets the F1 score $40\%$ on the test A set. Code is released at the \href{https://github.com/cybercore-co-ltd/aicity22-track4}{link}.
	\end{abstract}

	\section{Introduction}
	\label{sec:introduction}
\begin{figure}
	\centering
	\includegraphics[height=2cm, width=2cm]{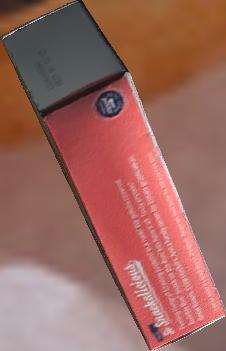} 
	\includegraphics[height=2cm, width=2cm]{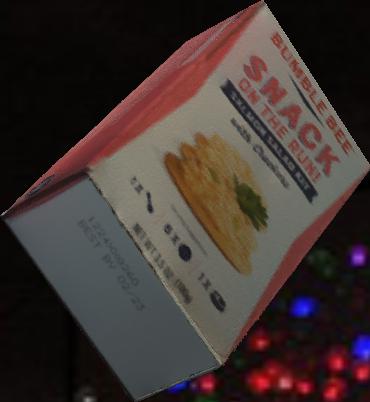} 
	\includegraphics[height=2cm, width=2cm]{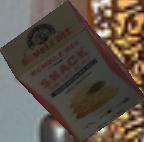} 
	\includegraphics[height=2cm, width=2cm]{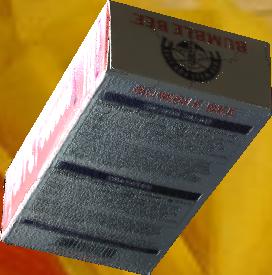}
	\includegraphics[height=4cm, width=\linewidth]{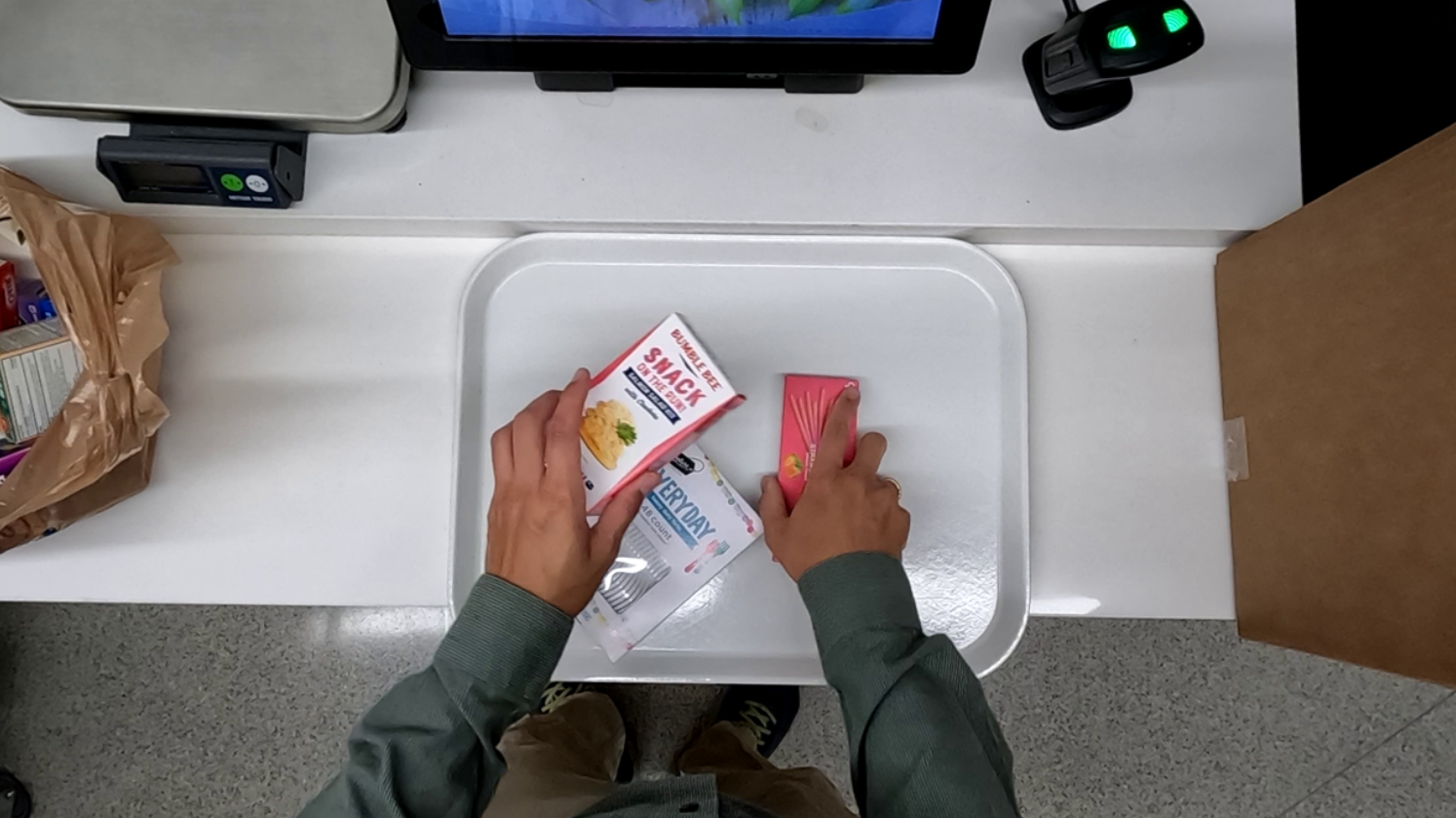}
	\caption{\small The top row is rendered images, and the bottom row is a test scene in AI City challenge 2022 -- Track 4.}
	\label{fig:2d_render}
\end{figure}

	Automatic checkout in retail stores is highly desired to relieve human labor and reduce mistakes. Computer vision techniques can potentially solve the problem by recognizing products to retrieve their price. However, a challenge is that items can appear under various poses and motions. Moreover, object overlapping and duplication also causes noisy input data. For these reasons, it is extremely time-consuming to collect rich dataset from real scene for training deep learning models. Therefore, synthetic data can be used instead. For example, in AI City challenge 2022 -- Track 4 \cite{Naphade22AIC22}, there are 116 product categories, each one is scanned to form a 3D model, then we can render their 2D images under numerous views. However, the appearance of synthetic samples is not similar to real ones, as illustrated in Fig. \ref{fig:2d_render}, which would raise a domain gap issue.

\begin{figure*}[!h]
	\centering
	\includegraphics[width=\textwidth]{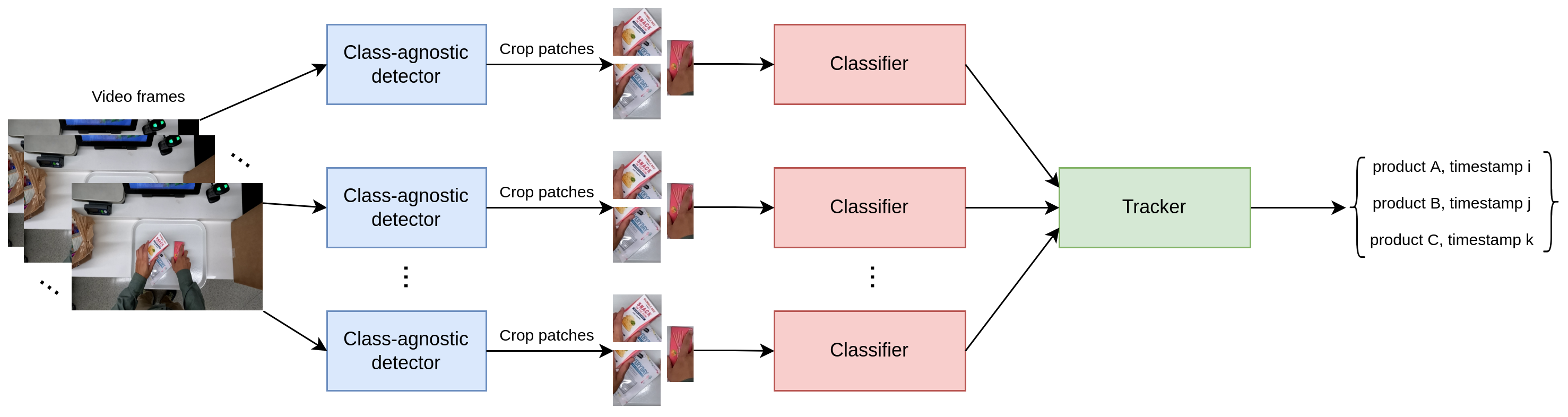}
	\caption{\small Our proposed pipeline. In the first stage, a class-agnostic detector proposes boxes for cropping products. Then, a classifier categorizes corresponding classes of patches. Lastly, a tracker link predictions across frames to form the final result.}
	\label{fig:pipeline}
\end{figure*}

In this work, we propose a solution for the Retail checkout. Its pipeline consists two stages: a detector for proposing product boxes and a classifier for categorizing product classes. Our method focuses on improving the domain generalization for the 1st-stage detector, and the main contribution is summarized as follows: \\
\begin{itemize}
	\item \textbf{Data generation} From rendered images, we construct a dataset to train detectors by replacing various background for the objects. Background can be plain color or synthesized by GAN models. While foreground objects are taken from the competition or ShapeNet \cite{shapenet2015}.
	\item \textbf{Learning without Forgetting} In preliminary experiments (Section \ref{sec:preliminary_experiments}), we found that COCO-pretrained detectors can sufficiently catch most of the objects. We hypothesize that these models are generalized enough for the task, therefore, we preserve this generalization during finetuning by Learning-without-Forgetting.
	\item \textbf{Ensemble} We ensemble different backbones to enhance the robustness of the 2nd-stage classifier.
\end{itemize}

The paper is organized as follows. Section \ref{sec:related_works} reviews the previous works, and Section \ref{sec:proposed_method} describes our main solution. Then, the experiments are presented in Section \ref{sec:experiments}, followed by conclusion in Section \ref{sec:conclusion}.

	\section{Related works}
	\label{sec:related_works}
	\subsection{Image classification}	
\label{sec:image_classification}
Since the benchmark paper AlexNet \cite{krizhevsky2012imagenet}, convolutional networks, such as VGG \cite{simonyan2014very}, ResNet \cite{he2016deep}, and Res2Net \cite{gao2019res2net}, has been the dominant approach for image classification. Recently, Transformer \cite{vaswani2017attention} attracts research attention with different variants, such as SASA \cite{ramachandran2019stand}, MLP-Mixer \cite{tolstikhin2021mlp}, ViT \cite{dosovitskiy2020image}, and Swin-Transformer \cite{liu2021swin}. The core success of Transformer is the attention mechanism. Specifically, long-range pixels can interact with each other to form a larger field of view, while convolutions can only capture local features. In our solution, we employ both convolution and transformer networks to diversify learned features.

\subsection{Object detection in Retail}	
\label{sec:object_detection}
Object detectors are mainly categorized into two approaches, i.e., two-stage and single-stage. Faster-RCNN \cite{ren2015} and Mask-RCNN \cite{he2017mask} stands for the two-stage approach, while RetinaNet \cite{lin2017focal}, ATSS \cite{zhang2020bridging}, and PAA \cite{kim2020probabilistic} are representatives for the one-stage approach. Applying to the retail, authors in \cite{goldman2019precise} created the SKU-110k dataset and used RetinaNet as the baseline. After that, \cite{yu2019solution} improved the result by using ATSS in conjunction with ensemble to boost the accuracy. Additionally, \cite{Ye2021ObjectDI} applied EM teacher \cite{he2020momentum} as self-distillation. Following previous methods, we adopt off-the-shelf object detectors pretrained on the MS-COCO dataset \cite{lin2014microsoft} to our task.

\subsection{Domain generalization}	
\label{sec:domain_generalization}

Domain gap occurs when training models on synthetic data and testing them on real one. This issue may make their behavior unpredictable. Some works \cite{zhou2021domain, tseng2020cross, zhou2020learning} are proposed to bridge the gap between the source and target domains. However, these methods require both source and target data during training, but it is not applicable in our case. To solve this restriction, we consider the Learning-without-Forgetting approach in \cite{li2017learning, serra2018overcoming, delange2021continual}. This method applies to a trained network on a dataset, then continues training the network on another dataset in such a way that the final network can perform accurately on both datasets.

	\section{Proposed method}		
	\label{sec:proposed_method}
	
	We propose a solution with a two-stage pipeline as illustrated in Fig. \ref{fig:pipeline}. First, an object detector proposes class-agnostic boxes for cropping products. Then, these patches are fed into a classifier to recognize corresponding categories. To avoid duplicated counting, we apply tracking and perform label voting. We also limit the region for tracking in the center of image. Finally, timestamp of a tracklet is estimated as the mean of the start and end time that the product is detected.

	\subsection{Preliminary experiments}
\label{sec:preliminary_experiments}

\begin{figure*}[!h]
	\centering
	\begin{tabular}{cccc}
		Color randomization & LSUN-GAN & CelebA-GAN & Big-GAN \\
		\includegraphics[width=0.2\linewidth, height=2.2cm]{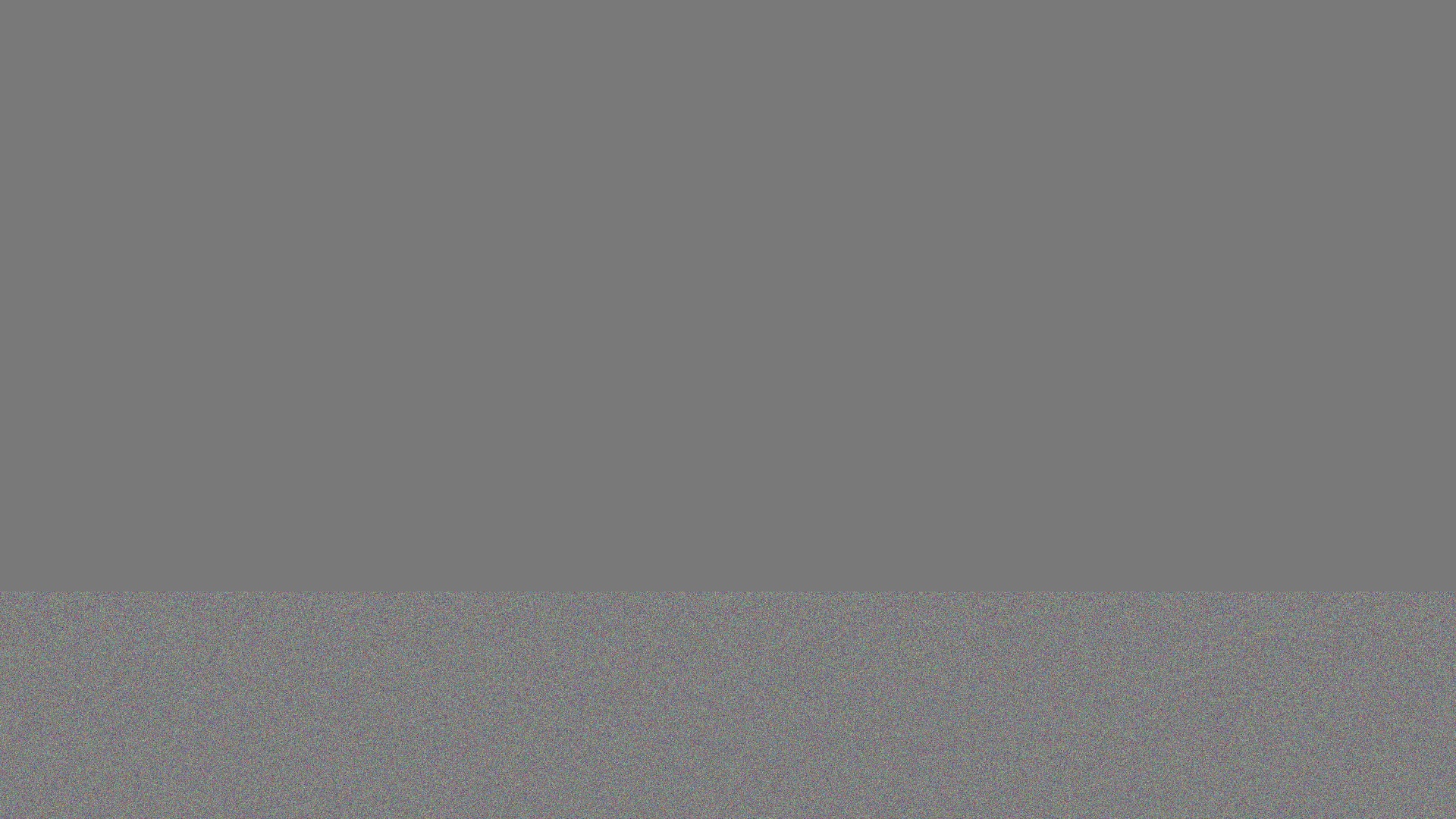} & \includegraphics[width=0.2\linewidth, height=2.2cm]{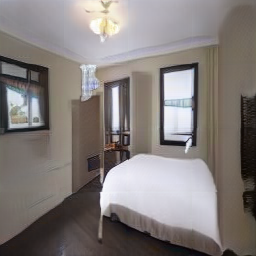} & \includegraphics[width=0.2\linewidth, height=2.2cm]{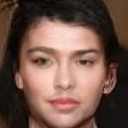} & \includegraphics[width=0.2\linewidth, height=2.2cm]{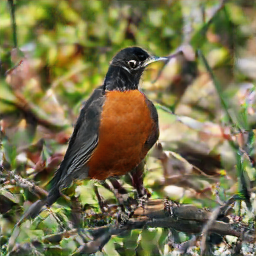} \\
		
		\includegraphics[width=0.2\linewidth, height=2.2cm]{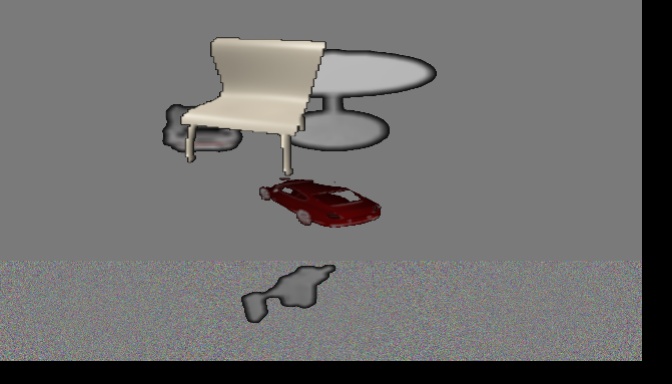} & \includegraphics[width=0.2\linewidth, height=2.2cm]{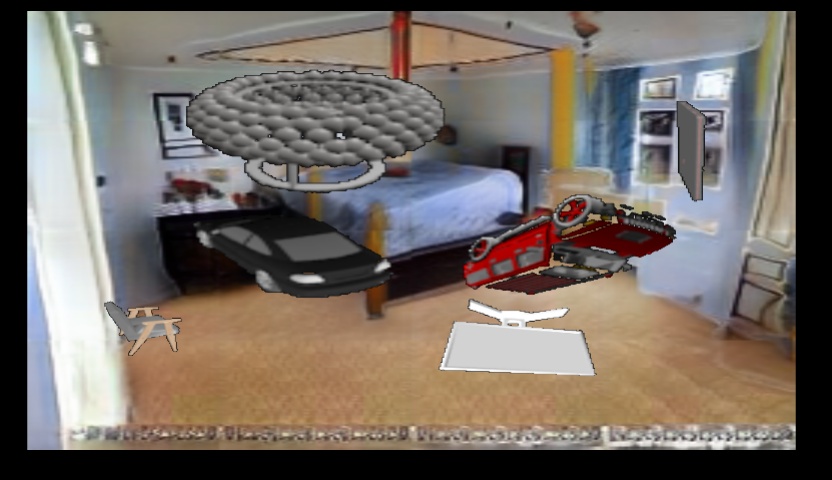} & \includegraphics[width=0.2\linewidth, height=2.2cm]{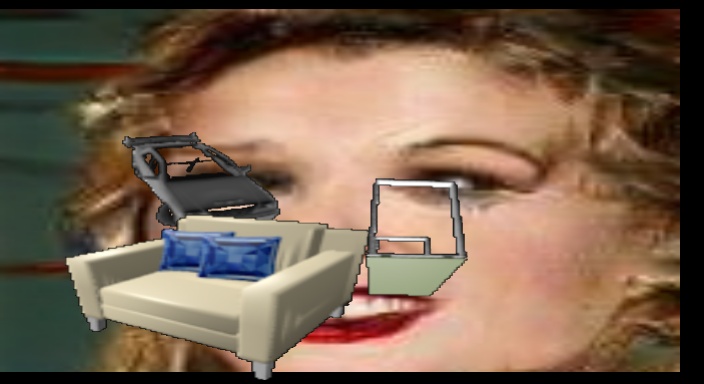} & \includegraphics[width=0.2\linewidth, height=2.2cm]{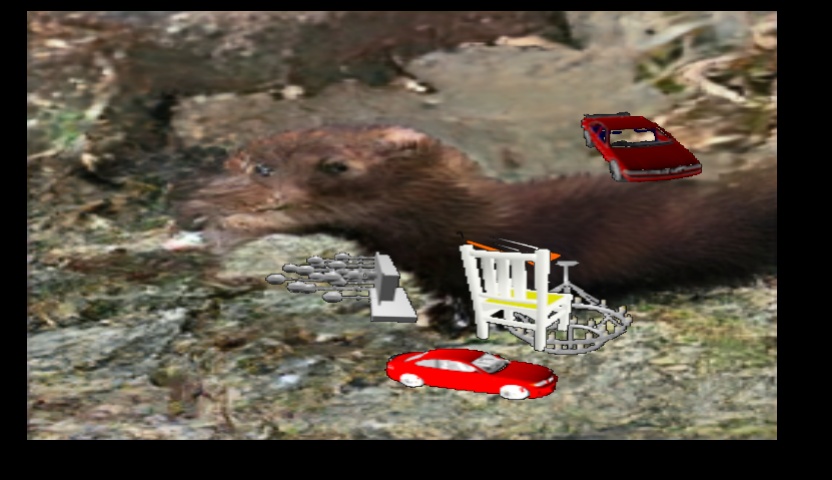} \\
		
		\includegraphics[width=0.2\linewidth, height=2.2cm]{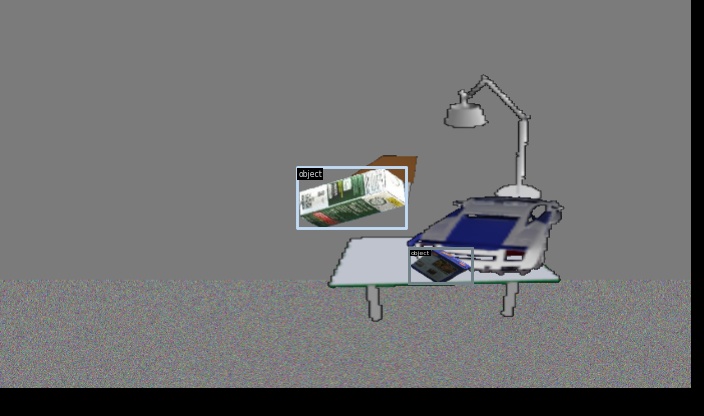} & \includegraphics[width=0.2\linewidth, height=2.2cm]{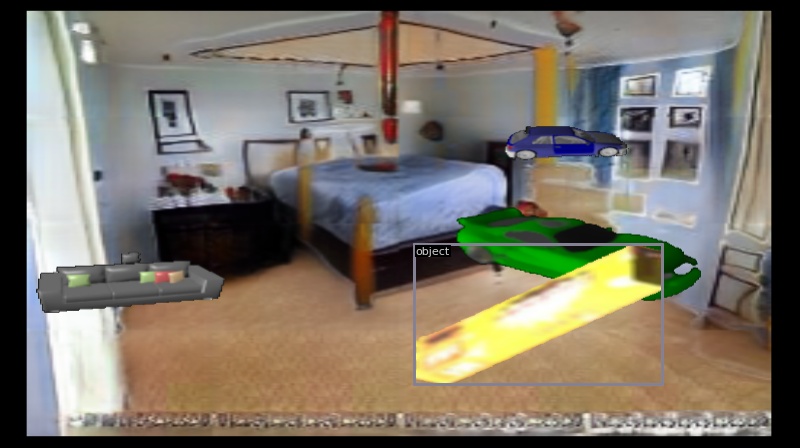} & \includegraphics[width=0.2\linewidth, height=2.2cm]{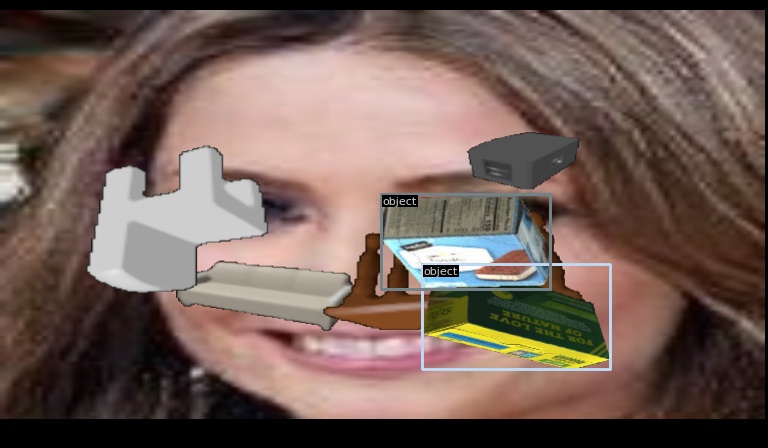} & \includegraphics[width=0.2\linewidth, height=2.2cm]{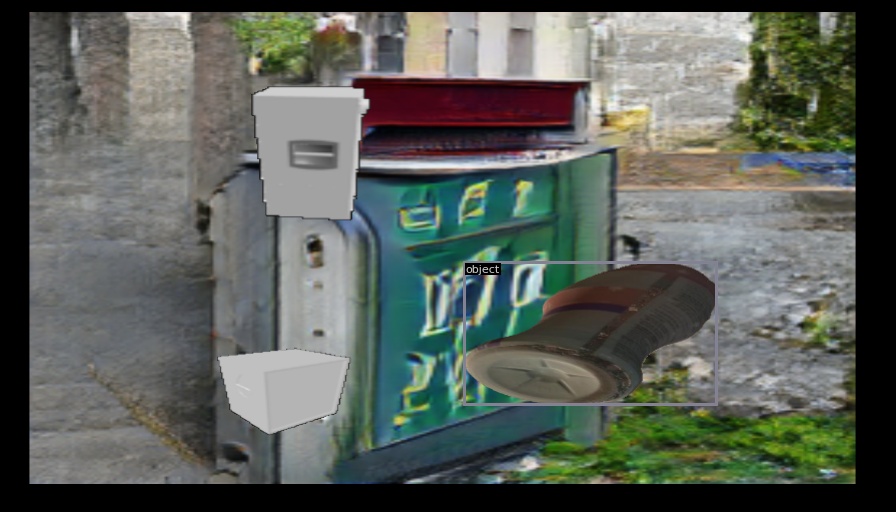} \\
	\end{tabular}
	\caption{\small Examples of datasets for training detectors. Columns are background generation methods, namely Color randomization, LSUN-GAN, CelebA-GAN, and Big-GAN. While rows are foreground sources, including No foreground, ShapeNet, and the competition.}
	\label{fig:data_gen}
\end{figure*}

We first generate a synthetic dataset, which is described in Section \ref{sec:data_generation}, for training detectors. In terms of classifiers, we utilize the dataset provided by the competition.

We compute the mean Average-Precision (mAP) score of the Mask-RCNN\footnote{\url{https://github.com/open-mmlab/mmdetection/blob/master/configs/swin/mask_rcnn_swin-s-p4-w7_fpn_fp16_ms-crop-3x_coco.py}} on our generated dataset and sort values to take the top-score classes. Then, we select 6 out of 80 classes, i.e., book, cup, bottle, hair drier, toothbrush, and remote, as useful classes. Afterward, we use boxes of these 6 classes to crop patches for the following classifier.

Regarding the classifier, we train the RepVGG-A0 \cite{ding2021repvgg} for 10 epochs and obtain $99.99\%$ top-1 accuracy. Finally, we cascade the detector and classifier to infer the test-A of the challenge and get $F1=0.2769$.
	\subsection{Data generation}
\label{sec:data_generation}

We employ background generation and foreground copy-paste. Concretely, the former consists of color randomization and synthesizing images by GAN models. Whereas, the latter copy objects from sources, including competition and ShapeNet datasets, then paste them into the background. The former is to diversify background concepts that help model robust to the real scene, whilst the latter is expected to reduce the false-positive rate.

\textbf{Color randomization} We randomly create plain-color scenes with value ranging between 120 and 220. The 1st column of Fig. \ref{fig:data_gen} shows several examples. 

\textbf{GAN} To make background more realistic, we use GAN models (see the 3rd and 4th columns in Fig. \ref{fig:data_gen}): COCO GAN \cite{lin2019cocogan} trained on LSUN and CelebA datasets, and BigGAN \cite{brock2018large} trained on ImageNet.

\textbf{ShapeNet} In our experiments, most COCO-pretrained detectors have a high false alarm rate. Hence, to improve their precision, we add objects from ShapeNet. Specifically, we randomly pick 2 to 6 objects to paste into synthetic backgrounds. Additionally, we apply several augmentation methods such as brightness adjustment, horizontal/vertical flip, and blurry, to make objects more diverse in appearance. The 2nd row of Fig. \ref{fig:data_gen} illustrate the examples. 
	\subsection{Learning without Forgetting}
\label{sec:learning_without_forgetting}

Image domain from testing videos is different from that of synthetic data, if we simply finetune pretrained models on this synthetic set, models will perform poorly on testing videos. Therefore, we adopt Learning Without Forgetting \cite{li2017learning} (LWOF), which was originally used for classification task, to our object detection problem. This method has the advantage that it can remember features learned from the old task when learning the new task. Another reason is that, we are not allowed to use any external data except synthetic one. Therefore, we cannot combine MS-COCO dataset with the synthetic dataset to finetune the model, this makes LWOF particularly suitable for this problem. Furthermore, LWOF acts as a regularizer to reduce the domain bias towards the synthetic data. 

In LWOF, we feed the same image to two models: teacher and student. The teacher is a pretrained model and frozen during training, while the student is initialized from the teacher, but contains two classification heads: the regularization head which is the same as teacher's head, and another is a classification head for learning the new task. Classification outputs from the teacher and student regularization head are enforced to be closed by a similarity loss. This ensures the model to not forget old learned features, whilst the student classification head is supervised by ground-truth label of the new task to learn the new patterns. Then, the teacher model and regularization head are removed after training. The Kullback-Leibler divergence loss is used to constrain similarity between teacher and student outputs:
\begin{equation}
L(y_{\text{st}},\ y_{\text{te}}) = y_{\text{st}} \cdot \log \frac{y_{\text{st}}}{y_{\text{te}}}
\end{equation}
 where $y_{\text{te}}$ is outputs from teacher, and $y_{\text{st}}$ is from student's regularization head.

\begin{figure}[]
	\centering
	\includegraphics[width=\linewidth]{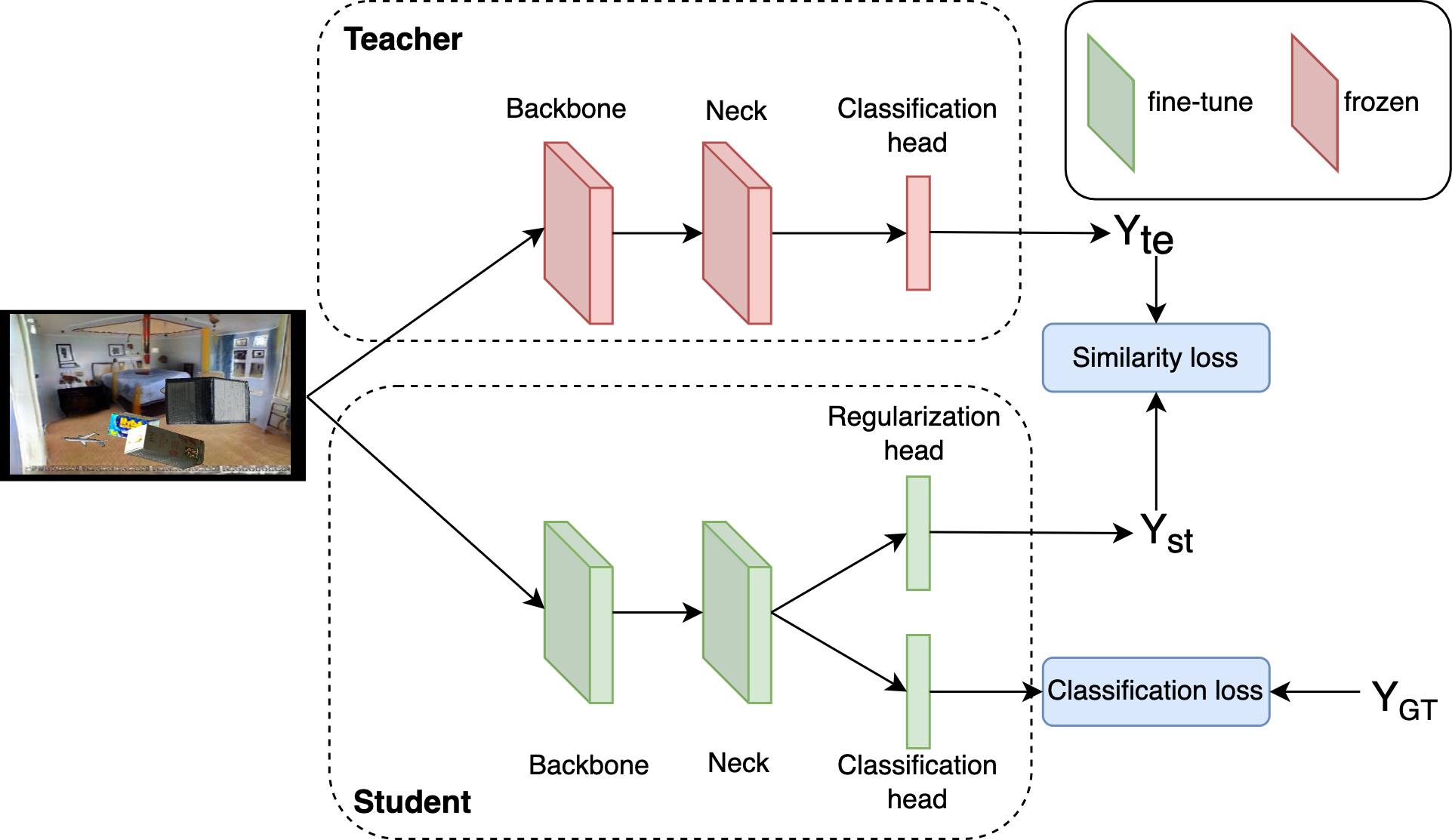}
	\caption{\small Illustration of the LWOF-style training for the detector.}
	\label{fig:lwof}
\end{figure}

Figure \ref{fig:lwof} illustrate how LWOF scheme works, where the mask head and regression head are omitted to highlight the key idea of LWOF. Classification outputs from the teacher's head and student's regularization head are involved in the similarity loss. Other components of the student model are kept as it is. We double the loss weight of the similarity loss to set the impact of preserving old learned feature higher. Finally, regression, mask, classification, and similarity losses are summed up.
	\subsection{Object detection Finetuning}
\label{sec:special-tricks}

In our first trials, we use COCO-pretrained models, and change the classification head from 80 classes to 1 class. Although models reach the high mAP on the val set, we found that these models are prone to be overfitted after 1 epoch. Since we can only use synthetic data, randomly initializing the classification weight may lead to being less generalized. Thus, while finetuning off-the-shell detectors, we keep the weight of the classification head unchanged (80 classes). To do this, for each ground truth, we duplicate ground truth 6 times corresponding to 6 classes listed in Section \ref{sec:preliminary_experiments}. Moreover, we empirically find out that using Sigmoid for the classification head is better than using Softmax. 
	\subsection{Post-processing}
\label{sec:post_processing}

\begin{itemize}
	
	\item \textbf{Ensemble} We apply Greedy Auto Ensemble \cite{liu20201st} with Weighted Box Fusion \cite{solovyev2021weighted} to fuse detection results of different models. In our experiments, we utilize model trained on COCO namely DetectoRS \cite{qiao2021detectors},  Swin Transformer \cite{liu2021swin}, TOOD \cite{feng2021tood}, Trident \cite{li2019scale}, Varifocalnet \cite{zhang2021varifocalnet}, and our finetuned one. For COCO-pretrained models, we only take 6 useful classes. Regarding classifiers, we average scores of models, e.g., RepVGG-A0 \cite{ding2021repvgg}, Swin-Transformer-S \cite{liu2021swin}, and Res2Net-50 \cite{gao2019res2net}.

	\item \textbf{Tracking} To avoid duplicated counting, we perform tracking to link detected objects across frames. We use ByteTrack \cite{zhang2021bytetrack} with IoU-score and Kalman-filter mode. Concretely, low-confident samples ($det\_score < 0.3$ or $cls\_score < 0.3$) are filtered out before feeding into the tracker, then boxes with IoU greater than $0.8$ are linked together. Finally, tracks that exist with less than $15$ frames are removed.
	
	\item \textbf{Label voting} After linking objects, label is decided for each track through label voting process. Assume a track $T=\{t_1, t_2, ..., t_N\}$ has $N$ items, in which, the item $t_i$ is categorized as class $l_i$. Then, we compute the contribution score $c_i$ for the item $t_i$:
	\begin{equation}
		c_i = f_{l_i}^{\alpha} * softmax( a_i^{\beta} * s_i^{\gamma} / \tau )
	\end{equation}
	where $f_{l_i}$ is the frequency that the class $l_i$ appears in the video, $a_i$ and $s_i$ are the bounding box area and detection score of the item $t_i$, and $\tau$ is the softmax temperature factor. Besides, $\alpha$, $\beta$, and $\gamma$ are tuneable hyper-parameters. Finally, the class $l_i$ corresponding to the highest contribution score $c_i$ in the track is selected to be the label representing the track.
	
\end{itemize}

	\section{Experiments}
	\label{sec:experiments}
	\subsection{Object detectors}
\label{sec:exp_object_detectors} 

\begin{table}[]
	\centering
	\small
	\begin{tabular}{c|c|cc|c}
		\hline
		Exp. & \multicolumn{1}{c|}{Detector}                            & Finetuning & LWOF & mAP  \\ \hline
		A1   & \multicolumn{1}{c|}{Mask-RCNN \cite{ren2015}}  			& \multicolumn{2}{c|}{}                  & 50.6 \\ \hline
		A2   & \multicolumn{1}{c|}{Mask-RCNN}                           & -          & -    & 47.6 \\ 
		A3   & \multicolumn{1}{c|}{Mask-RCNN}                           & \checkmark & -    & 66.0 \\ 
		A4   & \multicolumn{1}{c|}{Mask-RCNN}                           & \checkmark & \checkmark    & 68.8 \\ \hline
		A5   & \multicolumn{1}{c|}{DetectoRS \cite{qiao2021detectors}}  & \multicolumn{2}{c|}{}           & 67.0 \\ 
		A6   & \multicolumn{1}{c|}{TOOD \cite{feng2021tood}}       		& \multicolumn{2}{c|}{}           & 63.2 \\ 
		A7   & \multicolumn{1}{c|}{TridentNet \cite{li2019scale}} 		& \multicolumn{2}{c|}{}           & 52.6 \\ 
		A8   & \multicolumn{1}{c|}{VFNet \cite{zhang2021varifocalnet}}  & \multicolumn{2}{c|}{}           & 59.8 \\ \hline
		A9   & \multicolumn{3}{c|}{Ensemble A4, A5, A6, A7, A8}                                           & 78.6 \\ \hline
	\end{tabular}
	\caption{\small Validation of object detectors.}
	\label{tab:ob-result}
\end{table}

Our source code for object detection is based on MMDetection \cite{mmdetection}. All the detection results are reported in Tab. \ref{tab:ob-result}. We first use a COCO-pretrained Mask-RCNN with Swin-Transformer-S backbone to evaluate on the val set. We only keep 6 useful classes as described in Section \ref{sec:preliminary_experiments}. As a result, we get $50.6\%$ mAP (Exp. A1). Next, we finetune the model on the train set to see how much the model can be improved. Unfortunately, the result drops by $3\%$ mAP (Exp. A2). It is straightforward that the model seems to be affected by the domain difference. Then, in Exp. A3, we use the method presented in subsection \ref{sec:special-tricks} and get an improvement to $66\%$ mAP. Moreover, by training network with LWOF manner, the result reaches $68.8\%$ mAP.

We then make 4 experiments with models A5, A6, A7, and A8, trained on MS-COCO dataset without finetuning, then achieve results $67.0\%$, $63.2\%$, $52.6\%$, and $59.8\%$ mAP, respectively. Finally, we make an ensemble of 5 models in Exp. A9, obtaining $78.6\%$ mAP.

\subsection{Image classifiers}
\label{sec:exp_image_classifier} 
Our source code for image classification is based on MMClassification \cite{2020mmclassification}. We train 3 classifiers (10 epochs for each one), including RepVGG-A0 \cite{ding2021repvgg}, Swin-Transformer-S \cite{liu2021swin}, and Res2Net-50 \cite{gao2019res2net}. The results are shown in Tab. \ref{tab:classifier_validation}. All 3 models get high accuracy on the val set.

\begin{table}[!h]
	\centering
	\small
	\begin{tabular}{c|cc}
		\hline
		Model             & Top1 accuracy & Top5 accuracy \\ \hline
		RepVGG-A0         & 99.99\%       & 100.00\%      \\
		SwinTransformer-S & 99.82\%       & 99.99\%       \\
		Res2Net-50        & 99.98\%       & 100.00\%      \\ \hline
	\end{tabular}
\caption{\small Image classifiers validated on the val set}
\label{tab:classifier_validation}
\end{table}

\subsection{Two-stage pipeline}
\label{sec:exp_two_stage_pipeline} 

Experiments for the whole pipeline are listed in Tab. \ref{tab:pipeline_validation}. In Exp. B1, the combination of Mask-RCNN and RepVGG-A0 scores at $F1=0.2769$. Then, in Exp. B2, we ensemble 3 classifiers (RepVGG-A0, Swin-Transformer-S, and Res2Net-50) while keeping the same detector, resulting in $F1=0.3273$. In which, the precision is increased by about $6\%$ which indicates that the ensemble improves the classification robustness. Finally, we ensemble detectors and get an improvement to $F1=0.4000$ in Exp. B3.

\begin{table}[!h]
	\centering
	\small
	\begin{tabular}{c|cc|ccc}
		\hline
		Exp. & Detector  & Classifier & F1     & Precision & Recall \\ \hline
		B1   & Mask-RCNN & A0         & 0.2769 & 0.2045    & 0.4286 \\
		B2   & Mask-RCNN & Ensemble   & 0.3273 & 0.2647    & 0.4286 \\
		B3   & Ensemble  & Ensemble   & 0.4000 & 0.3448    & 0.4762 \\ \hline
	\end{tabular}
\caption{\small Two-stage pipeline evaluated on Test-A videos}
\label{tab:pipeline_validation}
\end{table}

	\section{Conclusion}
	\label{sec:conclusion}
	In this paper, we propose a solution for the automatic retail checkout by a two-stage pipeline, including class-agnostic detector and category classifier. We empirically find out that the COCO-pretrained weight is essential to detectors so we propose to finetune them with the learning-without-forgetting regularization so as to improve the domain generalization. Our solution scores at $F1=40\%$ on the test-A of the AI City challenge 2022 - Track 4.

\section*{Acknowledgement}
We would like to thank Chuong Nguyen and colleagues for valuable discussion and helping reviewing the paper.

	{\small
		\bibliographystyle{ieee_fullname}
		\bibliography{egbib.bib}
	}
\end{document}